# Classical Sorting Algorithms as a Model of Morphogenesis:
## self-sorting arrays reveal unexpected competencies in a minimal model of basal intelligence

**Authors:** Zhang, T.[2], Goldstein, A.[3,4], Levin, M.[1,2*]

**Affiliations:**
[1] Allen Discovery Center at Tufts University, Medford, MA 02155, USA.
[2] Wyss Institute for Biologically Inspired Engineering, Harvard University, Boston, MA 02115, USA.
[3] Department of Physiology, Anatomy and Genetics, University of Oxford, Oxford, UK
[4] Astonishing Labs.

*****Corresponding author:** michael.levin@tufts.edu

**Abstract**

The emerging field of Diverse Intelligence seeks to identify, formalize, and understand commonalities in behavioral competencies across a wide range of implementations. Especially interesting are simple systems that provide unexpected examples of memory, decision-making, or problem-solving in substrates that at first glance do not appear to be complex enough to implement such capabilities. We seek to develop tools to help understand the minimal requirements for such capabilities, and to learn to recognize and predict basal forms of intelligence in unconventional substrates. Here, we apply novel analyses to the behavior of classical sorting algorithms, short pieces of code which have been studied for many decades. To study these sorting algorithms as a model of biological morphogenesis and its competencies, we break two formerly-ubiquitous assumptions: top-down control (instead, showing how each element within a array of numbers can exert minimal agency and implement sorting policies from the bottom up), and fully reliable hardware (instead, allowing some of the elements to be "damaged" and fail to execute the algorithm). We quantitatively characterize sorting activity as the traversal of a problem space, showing that arrays of autonomous elements sort themselves more reliably and robustly than traditional implementations in the presence of errors. Moreover, we find the ability to temporarily reduce progress in order to navigate around a defect, and unexpected clustering behavior among the elements in chimeric arrays whose elements follow one of two different algorithms. The discovery of emergent problem-solving capacities in simple, familiar algorithms contributes a new perspective to the field of Diverse Intelligence, showing how basal forms of intelligence can emerge in simple systems without being explicitly encoded in their underlying mechanics.

# Introduction

On their respective time scales, evolutionary and developmental biology require that cognitive capabilities such as memory and goal-directed activity in the face of perturbations originate in proto-cognitive functions that existed long before complex brains came onto the scene [1-3]. The gradual history of intermediate forms with different levels of competency undermines a view in which discrete natural kinds have, or do not have, binary properties such as intelligence [3-8]. Moreover, a rich continuum of intermediate forms can be created by chimerizing biological and technological material in many different combinations [9, 10], further eroding the notion of a binary, categorical separation between engineered and biological capabilities. The nascent field of Diverse Intelligence seeks invariants across evolved, engineered, and hybrid systems to determine what all agents have in common, regardless of their composition or provenance, and thus better to understand the origin and scaling of embodied minds [11]. Beyond fundamental knowledge, this is also an essential step toward being able to recognize, repair, create, communicate with, and ethically relate to an enormous space of extant and forthcoming beings ranging from human cyborgs to synthetic life forms, Artificial Intelligences (AIs), and possible exobiological beings [10].

For the purposes of this study, "intelligence" refers (in William James' sense [1]) to some degree of competency in navigating a problem space so as to meet adaptive goals despite barriers, perturbations, and unexpected challenges. Human beings have much practice and skill in recognizing conventional intelligence; we easily detect it in the behavior of medium-sized objects moving at medium speeds in the 3-dimensional world. However, our evolutionary history, our outward-pointing sense organs, and our cognitive structure all make it difficult to detect unconventional intelligences that operate in novel embodiments, exist at different spatio-temporal scales, or live in unusual problem spaces [12]. In particular, while we have neuroscience and folk theory-of-mind for predicting the competencies of the collective intelligence of neural systems (i.e., animals with brains), we have no mature science that enables us to predict either the goals, or the degree of competency in pursuing those goals, of other kinds of systems.

Great strides have been made in understanding how complexity can emerge from simple local rules [13-17]. However, most of the emphasis to date has been on the emergence of phenomena at the lower end of degrees of the ladder of agency [18]. Beyond simple emergent complexity, such as that seen in static structures like fractals, lie second-order behaviors that serve as the origins of goals, preferences, valence, memory, and other phenomena that scale up to familiar cognitive systems [19-27]. Here, in keeping with an emphasis on basal (minimal) cognition, "goal" is used not to refer to a high-order, metacognitive "known purpose" as seen in human behavior, but rather in its minimal cybernetic [18] definition of a target state that a system has some ability to reach, despite a range of challenges. Learning to predict and control the goals of collective systems—especially newly engineered systems—is likely to be of existential importance to human flourishing over the coming decades in areas ranging from swarm robotics to AI systems to bioengineered tissues [28, 29].

Diverse Intelligence research includes the use of very minimal models to understand how problem-solving capacities can arise from the interaction of components at various levels [30]. These include synthetic droplets, molecular chemotaxis, and other simple systems in addition to the study of biological pathway models and whole cells and tissues [31-39].

One such system involves the collective behavior of cells during morphogenesis, such as embryonic development, regeneration, metamorphic remodeling, and cancer suppression [40-42].

What all of these phenomena have in common is the ability of cells (themselves composite, agential materials [43]) to work collectively to achieve specific anatomical endpoints [42], such as re-growing a limb and then stopping once the correct structure is complete [44], or remodeling a tail transplanted to the flank into a limb – the structure more appropriate to its new global location [45]. Crucially, this is not just open-loop emergence, but closed-loop control and error-minimization: cellular collective behavior in anatomical morphospace exhibits numerous abilities to solve problems (that is, to meet goals despite perturbations) in ways seen in other collective intelligences [42, 46]. For example, tadpoles which have the position of their facial features scrambled spontaneously re-arrange to the correct positions as they metamorphose into frogs [47, 48]. Examples abound of cells taking actions via molecular, cellular, and tissue-level actuators in order to reach a specific target state despite perturbations or changing circumstances (reviewed in [42]). Thus, it has been proposed that morphogenesis is the behavior of a collective intelligence of cells in anatomical morphospace, and that understanding and learning to exploit the problem-solving competencies of living tissues offers the opportunity for significant advances in regenerative medicine of cancer, traumatic injury, and birth defects [49, 50].

A large body of existing work explores the complex behavioral responses and capacities of tissues, cells [51, 52], and even of molecular pathways [31, 32, 53]. However, all of these systems and their internal subsystems are quite complex, exhibiting a likely endless variety of new details which could be responsible for the observed behaviors. To understand the necessary and sufficient dynamics for competencies to emerge requires insights from even more minimal systems — ones in which all of the components and their interactions are known and precisely trackable. Especially useful models for this research agenda leave no room to posit additional explicit mechanisms, or substrates encoding behavioral policies or goals, that have simply not been discovered yet. Thus, we are interested in toy models of collective decision-making that are entirely transparent, to gain insight into the lower bound at which unexpected behavior and problem-solving competencies can arise which may be relevant to cellular swarms.

Here, we abstract one key property of regulative morphogenesis: the ability to produce an anatomical structure with a precise order of components along one axis. For example, development or metamorphosis results in a tadpole or frog in which all of the organs are placed in a specific order along the anterior-posterior axis (Figure 1). For the purpose of modeling, we are agnostic as to whether this behavior occurs from scratch, such as during embryogenesis, or by unscrambling existing components (such as during metamorphosis and regenerative remodeling).

We abstract this task, undertaken by cells which can rearrange the organs as needed even when starting from highly abnormal initial configurations [48, 54], as a *sorting algorithm*. We use traditional sorting algorithms, as studied by countless computer science students, as a minimal system, and we study which unexpected, novel competencies these familiar algorithms might have in order to explore the idea that novel capabilities may lie in systems that we think we fully understand because we designed them.

To improve the fit between this model and the abilities of regulative development, we break two critical assumptions normally used with sorting algorithms. First, instead of a central algorithm operating on an array of numbers it can see and control in its entirety, we implement a distributed algorithm that is executed, in parallel, by each number (i.e., cell) with local knowledge of its environment. In lieu of a central controller, cells have individual preferences about the ordering between them and their neighbors. Second, we do not assume that each operation succeeds—that is, we (like biology) implement an unreliable substrate, in which some cells are defective and may not be able to obey when the rules tell them to move. We then quantitatively investigate the ability of these algorithms to sort a array of integers.

Our goals here are: 1) to establish a proof of concept for taking a system which seems simple and well-understood, and for using empirical experiments to identify that system's novel capabilities, goals, behaviors, and failure modes [55]; 2) to gain insight into the dynamical process of establishing a linear axis, so that the relevant dynamics could be better understood by developmental biologists and synthetic bioengineers [56-63]; 3) to understand how decentralized, agent-based systems can solve morphogenetic control tasks; 4) to determine how noise and unreliability in the medium is handled by such algorithms (robustness); and 5) to identify new behaviors and competencies that are not encoded overtly in the algorithm. Although ours is a very simple system, especially compared to any real biology, the benefit of these sorting algorithms is precisely that they are simple, easy to understand, and offer no place for additional complexity to hide (unlike in real cells). Here we show that even familiar, simple algorithms have the surprising ability to deal with perturbations in order to meet the algorithmically specified goals, and also exhibit novel behaviors that are not directly encoded in the algorithm.

# Definitions of Terms

| Term | Definition |
| --- | --- |
| Cell | Basic element to be sorted by the sorting algorithms. Each cell has an integer value property which is used for the comparison during the sorting process. |
| Cell-View Sorting Algorithm | Every cell follows an algorithm for making decisions as to how it can swap positions with neighbors to optimize local monotonicity of the integer values. We call it a cell-view algorithm because state evaluations and move decisions are made from the local perspective of each cell, rather than from the global perspective of a single, omniscient top-down controller. |
| Algotype | Each cell utilizes one of several sorting algorithms to dictate its movement. This "Algotype" is constant for the life of the cell (i.e. roughly equivalent to a fixed genetic or phenotypic identity). |
| Cell Value | The fixed integer value of each cell, which guides how it behaves in any of the algorithms (which are designed to order the cell values in sequence). |
| Experiment | Each sorting process starts from a randomized array of cells and runs until it meets the stop condition. |
| Active Cell | A cell that behaves normally during the sorting process as determined by the embedded cell-view sorting algorithm. |
| Frozen Cell | A cell that does not always move, even though the algorithm tells it to move (representing a damaged cell). There are two types. A "movable" Frozen Cell will not move on its own (will not initiate a move), but other cells are able to move it. An immovable Frozen Cell can neither proactively move itself nor can it be moved by another cell. |
| Delayed Gratification | The ability to temporarily go further away from a goal in order to achieve gains later in the process. None of the algorithms have this capability explicitly coded; where present, it is an emergent property of the algorithm dynamics. |
| Cell Aggregation | A metric of the degree to which the same type of cells cluster together (spatial proximity) during the sorting process when different Algotypes are mixed into the same sorting experiment. |
| Probe | The programming object which is created to monitor and record the status of the sorting process. |

## Methods

We developed a sorting algorithm evaluation system and implement the cell-view sorting algorithms in *python 3.0*. The following sections provide more details about the model of the sorting platform, the structure of the sorting cells, the process of the evaluation, and the experimental test methods. https://github.com/Zhangtaining/cell_research

Sorting Evaluation System

We designed the sorting evaluation system to consist of 2 parts: the sorting algorithm execution (which performs the sorting on a given array) and the sorting process evaluation (which oversees and analyzes the sorting across trials).
1. Sorting algorithm execution
   The sorting execution part chooses the specified sorting algorithm to perform the sorting experiments based on the given number of experiments and the Frozen Cells.
   The execution subsystem passes a Probe object to each experiment run, and the Probe is designed to record each step of the sorting process.
   After the sorting process ends, the information collected by the Probe is stored as a .npy file.
2. Sorting process evaluation
   The input for the evaluation is configurable including the algorithms to evaluate, the number of Frozen Cells, and the evaluation types. The evaluation process fetches the files based on the specified inputs. The evaluation subsystem picks up the corresponding files based on the inputs that contain the sorting process info. Then the given evaluation is performed for the data in those files.

Traditional Sorting Algorithms

In conventional sorting algorithms, a single top-down controller implements a set of rules to move cells around. The traditional algorithms we used as our baseline were:
- Bubble Sort
    1. Start at the beginning of the array.
    2. Compare the first two elements. If the first element is larger than the second, swap them.
    3. Move to the next pair of elements (second and third) and perform the same comparison and swap if needed.
    4. Continue this process, comparing and swapping adjacent elements as you move through the array.
    5. Each pass through the array will "bubble up" the largest unsorted element to its correct position at the end of the array.
    6. Repeat these steps for the remaining unsorted portion of the array until no more swaps are needed, indicating that the entire array is now sorted.

- Insertion Sort
    1. Start with the first element as the sorted portion (since a single element is always considered sorted).
    2. Take the first element from the unsorted portion and compare it with the elements in the sorted portion, moving from right to left.
    3. Insert the selected element into its correct position within the sorted portion by shifting larger elements to the right.
    4. Move to the next element in the unsorted portion
    5. Repeat steps 2 - 3 until the entire array is sorted.

- Selection Sort
    1. Start with the entire array as the unsorted portion.
    2. Find the smallest (or largest) element in the unsorted portion.
    3. Swap this element with the first element in the unsorted portion, effectively moving the smallest (or largest) element to its correct position in the sorted portion.
    4. Move the boundary between the sorted and unsorted portions one element to the right.
    5. Repeat steps 2-4 until the entire array is sorted.

Implementation of Cell-View Sorting Algorithm

We sought to study the sorting process in a more biologically grounded (distributed) architecture, where each cell is a competent agent implementing local policies. We thus defined three bottom-up versions of common sort algorithms, where actions take place based on the cells' perspective (view) of their environment within the array. We used multi-thread programming to implement the cell-view sorting algorithms. 2 types of threads were involved during the sorting process: cell threads are used to represent all cells, with each cell represented by a single thread; a main thread is used to activate all the threads and monitor the sorting process. The cell threads were multiple instances of the same sorting class (i.e., each cell had the same Algotype, which determined which of the sorting algorithms that cell used to guide its behavior). Inspired by the 3 traditional sorting algorithms described above, we designed 3 kinds of cell-view sorting algorithms (Figure 2):

- Cell-view Bubble Sort
    1. Each cell is able to view and swap with either its left or right neighbor.
    2. Active cell moves to the left if its value is smaller than that of its left neighbor, or active cell moves to the right if its value is bigger than that of its right neighbor.

- Cell-view Insertion Sort
    1. Each cell is able to view all cells to its left, and can swap only with its left neighbor.
    2. Active cell moves to the left if cells to the left have been sorted, and if the value of the active cell is smaller than that of its left neighbor.

- Cell-view Selection Sort
    1. Each cell has an ideal target position to which it wants to move. The initial value of the ideal position for all the cells is the most left position.
    2. Each cell can view and swap with the cell that currently occupies its ideal position.

3. If the value of the active cell is smaller than the value of the cell occupying the active cell's ideal target position, the active cell swaps places with that occupying cell.

Evaluation Metrics

To quantify the comparison between traditional sorting algorithms and their cell-view versions, we utilized the following metrics to evaluate the performance of those algorithms.

- Total Sorting Steps, Average and Standard Deviation
  We defined each comparison or swap as a sorting step, and we used the Probe to record the total number of sorting steps for each experiment. By comparing the average and standard deviation of the total steps, we derive the efficiency of sorting performance.

$$C = \frac{\sum_i^N c_i}{N}$$

$$\sigma = \sqrt{\frac{\sum_i^N (c_i - C)}{N}}$$

- Monotonicity and Monotonicity Error
  Monotonicity is the measurement of how well the cells followed monotonic order (either increasing or decreasing). The monotonicity error is the number of cells that violate the monotonic order and break the monotonicity of the cell array. The following formula shows the calculation for the monotonicity error for increasing order sequence.

$$E = \sum_i^N (i = 0 | V_i \geq V_{i-1} \rightarrow 0) \wedge (V_i < V_{i-1} \rightarrow 1)$$

- Sortedness Value
  Sortedness Value is defined as the percentage of cells that strictly follow the designated sort order (either increasing or decreasing). For example, if the array were completely sorted, the Sortedness Value would be 100%.

$$S = \frac{\sum_i^N (i = 0 | V_i > V_{i-1} \rightarrow 1) \wedge (\neg(i = 0 | V_i > V_{i-1}) \rightarrow 0)}{N}$$

- Sortedness Delayed Gratification
  Delayed Gratification is used to evaluate the ability of each algorithm undertake actions that temporarily increase Monotonicity Error in order to achieve gains later on. Delayed Gratification is defined as the improvement in Sortedness made by a temporarily error-increasing action. The total Sortedness change after a consecutive Sortedness value's increasing is $\Delta S_{increasing}$. The total Sortedness change after the consecutive Sortedness value decreasing starting from last peak is $\Delta S_{decreasing}$:

$$D = \frac{\Delta S_{increasing} - \Delta S_{decreasing}}{\Delta S_{decreasing}}$$

- Aggregation Value

In sorting experiments with mixed Algotypes, we measured the extent to which cells of the same Algotype aggregated together (spatially) within the array. We defined Aggregation Value as the percentage of cells with directly adjacent neighboring cells that were all the same Algotype.

$$A = \frac{\sum_{i}^{N}(T_i = T_{i-1} \rightarrow 1) \wedge (T_i \neq T_{i-1} \rightarrow 0)}{N}$$

Statistical Hypothesis Test Methods

We apply standard statistical hypothesis methods, Z-test and T-test, to evaluate the significance of the differences we report.

# Results

We first analyzed the results both from our Cell-View Sorting Algorithms and from the traditional versions of those sorting algorithms, with the goal of determining whether the cell-view versions worked (Figure 3), and comparing measures of efficiency, error tolerance, and Delayed Gratification with those of their canonical counterparts.

Efficiency comparison

We used the total steps that each algorithm needed to complete the sorting process for 100 elements in each experiment to indicate the efficiency of the algorithm (Figure 4). By repeating the experiments and doing the Z-test over the average steps, we calculated the efficiency difference between traditional sorting and cell-view sorting algorithms.

When we counted only swapping steps, the Z-test statistical values comparing the efficiencies of Bubble and Insertion sort were 0.73 and 1.26 (p-values were 0.47 and 0.24 respectively), revealing no significant difference between their performance. This indicates that the efficiency is very similar between traditional and cell-view versions of Bubble and Insertion sorting algorithms. However, cell-view Selection sort takes more swaps to complete sorting process than its traditional version by 11 times ($z=120.43$, $p<<0.01$). Thus, we conclude that the cell-view Selection sort is less efficient than the traditional Selection sort.

The situation changed when we considered both reading (comparison) and writing (swapping) as costly steps, simulating the metabolic cost of both measurements and actions. In this comparison, the total steps took to complete the sorting process of bottom-up vs. traditional sorting algorithms was decreased by 1.5 and 2.03 times for Bubble and Insertion sort respectively ($Z=-68.96$, $-71.19$, $p<<0.01$ in both cases). For Selection sort, the total steps of the bottom-up version were more, by 1.17 ($z=106.55$, $p<<0.01$). These results indicate that the cell-view Bubble and Insertion sorting algorithms are more efficient than the traditional versions. This is likely because traditional algorithms are using each element to compare with other elements, while cell-view algorithms will stop proactively comparing with other cells when they are on the target position. In contrast, the cell-view Selection sorting algorithm is less efficient than the traditional Selection sorting algorithm.

Error tolerance

To compare the error tolerance of the cell-view sorting algorithms with that of the traditional sorting algorithms, we introduced Frozen Cells into the sorting process. We ran the sorting experiments using different numbers of Frozen Cells and then checked the average final monotonicity errors for the experiments with a given number of Frozen Cells. A higher monotonicity error indicates lower error tolerance (Figure 5). We found that all the cell-view sorting algorithms exhibited less monotonicity error than the traditional versions, from which we conclude that cell-view algorithms have higher error tolerance than the traditional versions.

By comparing the different cell-view algorithms, we saw that with moveable Frozen Cells, the cell-view Bubble sort has the least monotonicity error (average value of 100 experiments was 0 with 1 Frozen Cell, 0.8 with 2 Frozen Cells and 2.64 with 3 Frozen Cells); and the cell-view Selection sort has the highest monotonicity error (average value of 100 experiments was 2.24 with 1 Frozen Cell, 4.36 with 2 Frozen Cells and 13.24 with 3 Frozen Cells). With immovable Frozen Cells, the cell-view Bubble sort has the highest monotonicity error (average value of 100 experiments was 1.91 with 1 Frozen Cell, 3.72 with 2 Frozen Cells and 5.37 with 3 Frozen Cells);

and the cell-view Selection sort has the lowest monotonicity error (average value of 100 experiments was 1.0 with 1 Frozen Cell, 1.96 with 2 Frozen Cells, and 2.91 with 3 Frozen Cells).In conclusion, we can see that both Cell-view and Traditional algorithms have performed error tolerance. The cell-view Selection sort had the highest Error Tolerance with immoveable Frozen Cells, and that the cell-view Bubble sort had higher Error Tolerance with moveable Frozen Cells.

Characterization of Delayed Gratification

Delayed Gratification (DG) is the ability to temporarily go further away from a goal to achieve gains later in the process (Figure 6). To compare the Delayed Gratification (DG) of cell-view algorithms and traditional algorithms, we performed experiments for the cell-view and traditional versions of the sorting algorithms with different Frozen Cell number, and calculating the Delayed Gratification based on the results (Figure 7). All the algorithm showed the ability of Delayed Gratification. The average Delayed Gratification difference between the Cell-view and Traditional Bubble sort was found to be 0.16 ($z=34.04$, $p \ll 0.01$). The difference between the Cell-view and Traditional Insertion sort was very small - 0.03 ($z=0.60$, $p=0.55$). The average Delayed Gratification difference between the Cell-view and Traditional Selection sort was 2.77 ($z=17.21$, $p \ll 0.01$). From this we conclude that cell-view Bubble sort performs more DG than the traditional version, cell-view Insertion sort performs very similar amounts of DG as the traditional version, and cell-view Selection sort performs less DG than the traditional version.

A random walker will sometimes move further from its goal and thus exhibit what may look like Delayed Gratification. To demonstrate DG as a problem-solving strategy it must be shown to be performed specifically in the context of barriers, not just part of a random strategy. Thus, we next compared the amount of DG observed for each algorithm in the context of different numbers of Frozen Cells: would the algorithm tend to back-track in Sortedness *more when there are more broken cells in its environment?* We observed a clear trend of increasing average Delayed Gratification for the Bubble and Insertion sort experiments for both Traditional and Cell-view, and the Cell-view algorithms performedmore Delayed Gratification during the sorting process. The average Delayed Gratification for cell-view Bubble sort was 0.24 with 0 Frozen Cell, 0.29 with 1 Frozen Cell, 0.32 with 2 Frozen Cells, and 0.37 with 3 Frozen Cells (all average values are based on 100 repetitions). For Insertion sort, we saw that the average DG value was 1.1 with no Frozen Cell, 1.13 with 1 Frozen Cell, 1.15 with 2 Frozen Cells, and 1.19 with 3 Frozen Cells. However, we did not see a clear trend for either cell-view or traditional Selection sort. This reveals that Bubble and Insertion sort deploy Delayed Gratification in a context-sensitive manner – they do more backtracking when faced with defective cells and are not randomly back-tracking through their space.

Mixed Algotype sorting: analyzing chimeric arrays

We next introduce the notion of an "Algotype": this refers to one of several discrete algorithms that a cell may be using to control its behavior. Algotype is meant to be distinct from data quantities like a cell's numerical value or its current position; rather, Algotype reflects a cell's behavioral tendencies. Our use of bottom-up (distributed) control in the sort process allowed a new kind of experiment: a chimeric array in which different cells use different policies to achieve their objectives, analogous to biological experiments in which cells with different genetics or cell types were mixed in the same body (reviewed in [9]). We wondered: would chimeric arrays still self-sort, and what would be the behaviors of individual cells when their neighbors were using different algorithms? Note that cells had no explicit provision for detecting their own or their neighbors

Algotype explicitly; because the Algotype is a meta-property not addressed in any way in the algorithm itself, its consequences only become evident through the cells' behavior over time.

At the beginning of these experiments, we randomly assigned one of the three different Algotypes to each of the cells, and began the sort as previously, allowing all the cells to move based on their Algotype (i.e. their individual sorting algorithm). The process was considered to be completed when the Sortedness Value of the array stopped changing for several time steps.

The first observation from these experiments was that all of the Algotype combinations can completely sort the array (Figure 8A, blue lines), demonstrating that components with different policies but the same goal can be mixed in the same collective without abrogating the ability to complete the system-level task.

We next checked the efficiency: do chimeric arrays function as efficiently as homogenous ones? We compared the number of swapping steps in mixed-Algotype experiments with those required to reach sorted state in experiments with a single Algotype. The average steps to complete a pure cell-view Bubble sort was 2448.8 (n=100 replicates). The average steps to complete a pure cell-view Insertion sort was around 2482.8 (n=100 replicates). The average steps for to complete a pure cell-view Selection sort was around 1095.5 (n=100 replicates). For sorting with mixed Algotypes, the Bubble-Insertion mix took an average of 2476.02 steps to complete the sort (n=100 replicates). The Bubble-Selection mix took an average of 1740.9 steps to complete the sort (n=100 replicates). The Insertion-Selection mix took an average of 1534.77 steps to complete the sort (n=100 replicates). This revealed (Figure 8B) that the total number of steps used by mixed Algotypes sorts falls between the pure Algotype sort that takes the most steps and the pure Algotype sort that takes the fewest steps. Therefore, we conclude that the efficiency of a chimeric array is roughly the average of the efficiencies of its two component Algotypes (the efficiencies combine linearly).

We next looked for unexpected behaviors at the level of individual cells and groups of cells (corresponding to chimeric tissues within an organism) by examining the spatial location of cells with different Algotypes. Specifically, we computed the tendency of individual Algotypes to cluster together within the array as a way to determine if the Algotype has any influence on *how* those cells travel throughout the morphological space [12, 64, 65] making up the array structure, during the array's travel through its sorting space. To check whether the same Algotypes tended to gather closely together during the sorting process (Figure 8C), we tracked the position of different Algotypes and repeated the experiment multiple times to check the average Aggregation Value during the sorting process. It is important to note that this is not something that can be predicted *a priori* since none of the algorithms access the Algotype information of their own or neighboring cells—there are no explicit steps implementing clustering or any other kind of distribution.

As a negative control, we first ran the experiments using Algotypes that were in fact identical algorithms, as a sanity check to exclude the presence of irrelevant factors in the code (Figure 8B, light pink line). We found that the peak Aggregation Value of the Bubble-Insertion mix was 0.61 (std dev 0.04, N=100), the peak Aggregation Value of the Bubble-Selection mix was 0.65 (std dev 0.05, N=100), and the peak Aggregation Value of the Insertion-Selection mix was 0.57 (std dev 0.04, N=100) (Figure 8A), revealing the baseline that corresponds to no significant aggregation (just random assortment) among arbitrary elements following precisely the same algorithm.

However, when we analyzed the aggregation values in arrays with chimeric Algotypes (consisting of cells using distinct algorithms to guide their behavior), we observed a remarkable and unexpected effect (dark red lines in Figure 8A). At the beginning, the aggregation was 0.5 in all cases, as befits the random assignment of Algotypes to cells. At the end, they were also 0.5 because the final state is a fully-ordered array, and the random assignment of Algotypes to initial values means that is impossible to maintain non-random Algotype assortment when sorting on the cells' values (which are randomly related to their Algotype). However, during the sorting process itself, we observed a significant ($p<<0.01$) difference of the aggregation values from the negative control: distinct Algotypes cluster together, to maximal levels of 0.72, 0.65, 0.69 and 0.62 in experiments mixing Bubble and Selection; Bubble and Insertion; Selection and Insertion; and all three respectively. The maximum segregation occurs at 42, 21, 19 and 22 percent of the way through the process for Bubble and Selection; Bubble and Insertion; Selection and Insertion; and all three respectively. We conclude that the different Algotypes exhibit aggregation (spatial clustering) during the sorting process even though the cells have no way to directly read each other's type and none of the algorithms refer to that property explicitly.

Why does this happen? Based on the above data on algorithm efficiencies, we first hypothesized that cells with a more efficient Algotype would move to the desired positions first, and as the sort continues, the cells with less efficient Algotypes would "catch up" and move the aggregated cells to their final position, splitting up the initial clusters. This hypothesis predicts that the aggregation is entirely due to differences in the Algotypes' efficiency. On the other hand, there could be a more general phenomenon at play. To analyze this, we performed similar experiments as above but allowed assignment of duplicated values to cells (100 cells with values ranging from 1 to 10, guaranteeing duplicated occurrences of 10 cells for each value randomly distributed in the initial string). Thus, there was no explicit reason for one cell of value "5" (for example) to appear before or after another cell of value "5" by the time the whole array was sorted. In other words, this version of the experiment allows any clustering to be maintained through to the end of the sorting process because a set of numbers could now be at the correct position in terms of their value and yet be arranged in any degree of clustering according to Algotype within that region (Figure 8E).

We observed that the Aggregation Values rose and did not decrease for the Bubble-Selection mix and the Insertion-Selection chimeric arrays. The average final Aggregation Values for Bubble-Selection and Insertion-Selection mixes were 0.65 and 0.7 (repeated 100 times), which is higher than the highest Aggregation Values of Bubble-Selection and Insertion-Selection mixes with non-duplicate values. This suggests that the efficiency hypothesis can be the explanation for the segregation, because the Selection algorithm is more efficient than Bubble and Insertion while the efficiency is similar between Bubble and Insertion based on our previous analysis.

The experiments in which repeated cell values are allowed also enabled us to ask another question. If we release the pressure of needing to be numerically sorted at the end, how high would the aggregation go? In the unique-value experiments, cells clustered with their same Algotype inevitably eventually get pulled apart at the end to establish the correct final sort order. But, if there were multiple versions of each value with different Algotypes, they could remain next to each other while the whole cluster was in its numerically-proper position. Thus, we could see how high the natural tendency for emergent aggregation is, when not artificially limited by the explicit algorithm's need to sort the values. We performed these experiments (Figure 8D) and observed maximal levels of 0.69, 0.63, and 0.71 in experiments mixing Bubble and Selection, Bubble and Insertion, Selection and Insertion respectively. The maximum segregation occurred at 100, 13, and 100 percent of the way through the process for Bubble and Selection, Bubble and Insertion, Selection. This illustrates how the explicit goals of a mechanism, and its emergent behaviors, can be tested separately, and shows the ability for these emergent behaviors to aim for a specific parameter value (e.g., a segregation value of 0.69, not simply "maximize segregation").

The availability of chimeric arrays gave us one more interesting opportunity: what happens when the two different Algotypes are at cross-purposes—that is, they do not have the same goal? This corresponds to biological problems such as chimeras made of animals with different target morphologies [9, 66]: what will the cells end up building? To test this in our model, we performed experiments using two mixed Algotypes, where one was made to sort in *decreasing* order while the other sorted in *increasing* order. We ensured that all 3 combinations start from similar Sortedness, ~50%. At the end of the sorting process, none of them reached 100% sorted; instead, they ended with Sortedness values of 42.5, 73.73 and 38.31 in experiments mixing a) Bubble sorting decreasingly and Selection sorting increasingly, b) Bubble sorting increasingly and Insertion sorting decreasingly, and c) Selection sorting decreasingly and Insertion sorting increasingly, respectively. We observed (Figure 9) that the sorting trajectories of these 3 combinations were very different. For Bubble and Selection, the Sortedness dropped sharply at the beginning of the sorting process, then increased for a short period. After the Sortedness value reached around 48, it dropped again and stopped at the value below 44 For Bubble and Insertion, the Sortedness value monotonically increased and stopped above 50, while for Selection and Insertion the Sortedness value almost monotonically decreased during the sorting process and stopped below 50. Thus, the different Algotypes battle each other for some time but eventually reach a global equilibrium. This reveals how competition among chimeric subunits with different emergent local goals can be studied in the context of a collective system with explicit algorithmic goals, and the overall stable states that can be achieved.

**Discussion**

Morphogenesis - the self-assembly of complex anatomies during development or regeneration - can be understood as collective behavior of cells traversing morphospace [12, 42]. We constructed a simplified model of how cells or organs sort themselves along an axis during regulative development and regeneration [47, 48, 54] as being functionally similar to the task of sorting numbers along a number line. Organisms perturbed during development (for example by moving cells out-of-place), or dissociated cells allowed to re-aggregate, often find their way to the specific target morphology [55]. We used conventional sorting algorithms as a very minimal component of this repair process, and looked for behaviors from those algorithms that might be new to students of computer science who routinely utilize those algorithms. This strategy is part of the field of Diverse Intelligence in that it helps to recalibrate our intuitions on the complexity of underlying mechanisms that may be sufficient for basal competencies normally expected of advanced or even neuronally-based systems, and also helps us see navigation of diverse kinds of problem spaces as bona fide *behavior* that can be probed via interventional strategies.

We analyzed three such algorithms in their classical form, as well as in a new implementation where we discarded two ubiquitous assumptions in favor of more bio-realistic scenarios. First, instead of top-down algorithms that control an entire process as a single agent with a single algorithm (behavioral policy), we implemented the same algorithms from the cell's-eye view, as a local, distributed system in which each cell has preferences for what neighbors it will have, and has some capacity to move around in order to implement those preferences [67]. While biology does have important examples of global control and top-down signaling (reviewed in [68, 69]), many biological outcomes result from the activities of distributed, local agents. This perspective is implemented in many examples [58, 70-75] of using agent-based modeling in biology (although the actual agency possessed by those components is typically assumed to be low).

Second, we introduced the notion of damaged or malfunctioning cells that cannot move even when the algorithm says they should. While unreliable computing is an existing field [76, 77], the standard study of sorting algorithms assumes that the steps are followed correctly. By contrast, biology excels at managing a highly noisy, unreliable medium at every scale [78, 79], utilizing this condition as an important aspect of evolving problem-solving machines (not unique solutions) [41]. In biology, the noisiness of the local environment is a feature, not a bug, leading to a ratchet of continuously increasing competency when there is a developmental abstraction layer between genotypic specification of the hardware and the behavioral/functional phenotypes under selection [80-83].

We evaluated the Sortedness of the input array as a measure of the traversal of the algorithm through its problem space, and performed experiments which illustrate how even minimal, deterministic systems can be tested for novel behavioral competencies. First, we found that a cell-level version of algorithm does, in fact, work: it completes the task. When counting the cost of both taking measurements and making moves (for each cell), two of the three cell-view algorithms are actually significantly more efficient as distributed agents than as classical top-down algorithms.

Next, we assayed an important aspect of intelligence: the ability to, when faced with a Frozen Cell, go around it in a manner that temporarily takes one further from the goal (Delayed Gratification) [1, 12]. Pure strategies that seek to minimize error (like the magnets separated by a small piece of wood in William James' example [1]) cannot do this, and different creatures have different tolerances for this way of navigating the problem landscape. The possibility of this behavior in sorting algorithms has not to our knowledge been considered, possibly because in their traditional implementation, there are no Frozen Cells for them to go around. Here, we included "damaged cells", which were in effect a Frozen Cell in sorting space – the algorithm simply could not move a cell when it needed to, to continue its trajectory. We found that when their path in sort space is analyzed, it is seen that cell-view sorting algorithms do indeed exhibit this rerouting behavior, temporarily allowing Sortedness to decrease (moving away from their goal) in order to find a new set of steps that would the problem and ultimately improve monotonicity. This was shown to be context-dependent; that is, Delayed Gratification was a function of the number of Frozen Cells.

This is especially significant because our algorithms are deterministic; there was no stochasticity in the algorithms, as might typically be assumed to be needed to avoid local minima. In addition, our cell-view algorithms contained no explicit steps for what to do in case of a disobeying cell, or even any steps for assaying whether any of their actions have had the desired effect in the first place. In other words, our cells' algorithms were purely open-loop with no feedback. The fact that our systems nonetheless exhibited implicit goal-seeking with Delayed Gratification highlights two paths to robust biological goal-seeking: stochasticity *or* distributed components with fault-tolerance. While our simulations here only studied the latter, in future work we plan to investigate how behavior changes when both techniques can be utilized simultaneously.

We think it remarkable that these simple algorithms have the capacity to solve unexpected problems in their space, given that it is not explicitly encoded in the algorithm itself, which has no "metacognitive" steps that monitor the sorting progress. Indeed, the lack of explicit algorithmic steps to monitor improvements in Sortedness may well be a key benefit, because otherwise the algorithm may not have been allowed wander backwards when appropriate (i.e., in some cases, leaving explicit control out of the algorithm may enable greater problem-solving capacity). We believe these results, like the maze navigation seen by microbial cells and even simple chemical droplets [84-89], imply the need for experimental exploration even of systems whose simple, transparent nature can lull us into a false sense that we understand their capabilities just because we understand (or even created) their parts. This goes beyond first-order (static) emergent complexity (e.g., fractals or cellular automata) and draws attention to a next level: emergent behavioral competencies and the beginnings of minimal agency.

Lastly, we note that the use of bottom-up distributed algorithms allows the testing of something that is impossible in the classic version: a chimeric scenario in which some cells utilize a different sort algorithm than other cells. Chimerism at different scales [9] is often used in biology to address the role of the genome in the collective decision-making that determines large-scale form and function. Yet despite much progress on the molecular genetics guiding individual cell properties, the field generally has no models that make predictions on morphological outcomes from chimeric experiments in development or regeneration. That is, while we have a good understanding of the molecular hardware, we do not have good frameworks for understanding the collective decision-making of cells with different policies ("biological algorithms" for systems with multiscale competency) with respect to how they will navigate anatomical space. Simple models like ours which map the behavioral competencies of chimeric systems onto large-scale problem-solving are important for developing an understanding of the principles involved.

To begin to address this issue, we define the notion of the Algotype, referring to the behaviors of a given cell under various circumstances. We intend this concept to be significantly different from genotype (specification of the explicitly observable hardware) and from phenotype (outcomes that can be detected in a single observation, such as geometric or physiological states and properties). Note that in denoting the time-extended "personality" of cells, the Algotype consists of two components: the expected behavior overtly encoded in the algorithm (the sorting), and possible emergent tendencies not obvious from the mechanics and not explicitly assigned (e.g., the clustering). Thus, Algotypes include behavioral tendencies (such as preferring to associate with their "kin" during their journey through morphospace) that may not be explicitly encoded anywhere in the algorithm and can only observed as a holistic dataset encompassing many scenarios and situations [90]. In biology, multiple Algotypes can be instantiated by the same genetics, and multiple different genomes can result in hardware with similar Algotypes. We believe this notion will be useful to understand plasticity and scaling of competencies in biology [8, 12, 21, 46, 91-97] far beyond the toy model of multi-scale problem-solving explored here.

We found that chimeric arrays consisting of distinct Algotypes still manage to get sorted; thus, there is no need for each cell to be following the same algorithm as long as they have the same goal. However, if they are set to cross-purposes (one sorts for increasing, and the other sorts for decreasing), they reach a dynamic equilibrium at a mixed, intermediate state in which the large-scale demographics of the array (with respect to Algotype) no longer changes, but the individual cells can still move (akin to extensive biological turnover at the molecular and cellular scale while the large-scale anatomy is maintained over years).

This is, as far as we know, the only model that implies a prediction on what will happen when, for example, different species' stem cells (neoblasts) are mixed within one flatworm body. Given that each set of stem cells knows how to undergo division and metamorphosis until a specific head shape is completed, it is entirely unclear what will happen in a chimeric worm: will one type be dominant over the other, or will an intermediate shape result, or will there be endless remodeling because neither set of cells ever reaches its stop condition with respect to the morphology? Our model showed that despite no explicit affordance made for this scenario in the algorithm, the sort process included many islands within each array where cells of similar Algotypes clustered together within the space of the overall array. This suggests the testable hypothesis that what will occur in a chimeric biological body is the establishment of a set of tissue-level islands, each of which has the identity of one of the parent species (a patchwork of local neighborhoods that possess a tissue-wide but not organism-wide identity). Because our model did not include explicit mechanisms for cells to behave differently with neighbors of different Algotypes, we suggest that this biological clustering should occur even in the absence of, for example, distinct adhesion molecules present on the cells of different genomes or any other hardware-encoded mechanisms for cell sorting.

Finally, we found a most unexpected behavior in these chimeras. During the sorting process, cells with similar Algotypes (and thus the same sorting behaviors) tended to aggregate together spatially within the array. Eventually they are pulled apart by the necessity to sort themselves according to their positional (numerical) value (rather than by algorithm type), but before that happens, they act as though they have a strong affinity for each other despite the fact that the algorithm says nothing about aggregating, and has no explicit provision for a cell determining the Algotype of any given neighbor (or itself). We conclude that even very simple algorithms can have novel behaviors that are not explicitly encoded in their formal policies and are extremely hard to predict ahead of time. We also suggest that this simple model can be used to ask basic questions about large-scale outcomes in chimeric systems in which components have different explicit and implicit Algotypes. These kinds of analyses can begin to provide principled answers to currently open questions in biology, such as how to predict the morphogenetic outcomes in cellular chimeras consisting of different species' cells or engineered components that extend functional reach outside the standard body envelops [9, 10, 66, 98].

By allowing repeated digits, we were able to partially dissociate the pressures of the explicit algorithm (to sort elements based on numerical value) from the tendencies of the emergent aspect of the Algotype (to cluster with like-minded elements within the string). Existing real-world biological examples of releasing sub-agents from explicit control, to ask what they would do if allowed, include the engineering of biobots. It has recently been shown that un-modified, genetically-normal cells self-assemble into constructs with novel behaviors when freed from the instructive influence of their neighbors [28, 99], revealing baseline competencies not apparent from their standard role within default developmental algorithms. The minimal model shown here represents a first step toward the development of more general strategies to study emergent goals in collective systems and ways in which those goals cooperate with, compete with, and alter the performance of explicit goals we (or evolution) instantiate via hardware or software mechanisms.

There is no magic here, of course: everything that happens is, in some way, a consequence of the rules being followed. In the same way, truly cognitive behavior of living systems must be consistent with the physics of their smallest components. However, while such behaviors do not contradict the laws of physics of their world and can be explained after they are observed, they often require a different set of conceptual tools to effectively predict and exploit them than physics and chemistry [21, 30, 94, 96, 100-106]. We believe advances in our understanding will result from when we will be able to identify and predict *in advance* the preferences and behavioral competencies of novel collective systems we engineer. Note also that we observed not only capabilities related to the solving of the problem which the algorithm explicitly addressed (sorting numbers) but new behaviors (clustering) that do not seem related to the intended purpose for which their algorithm was written. This aspect goes beyond the well-known realization that simple rules can give rise to unexpected emergent behaviors, emphasizing the need to be able to predict and control the properties that such novel behaviors actively seek to maximize.

This model can be expanded in numerous ways. For example, what happens with cells that are not permanently broken, but have the ability to unfreeze given specific (or merely repeated) nudges by their neighbors? Also, we plan to investigate how general these findings are to algorithms for 2-dimensional ordering problems and others. Another interesting direction would be to mix distributed and top-down controls—to monitor the behavior of a classical algorithm working on an agential material that also moves in ways that it was not instructed to. This is the situation in biology, and the evolutionary and cognitive impacts of an agential medium are just beginning to be understood [41, 43]. Understanding what changes should be made to a top-down algorithm to facilitate and exploit the competencies of its medium would, for example, assist the design of multiscale robotics.

One limitation of the current analysis is that we looked at only one emergent behavior (aggregation); other interesting things could be happening that we do not yet know to test for, even in this system. More broadly, we believe it is necessary to develop frameworks for looking for novel competencies in systems, with an increased emphasis on broad, unbiased analyses to help human scientists find those goal-directed behaviors which our cognitive biases do not readily facilitate.

We suggest that the study of these kinds of dynamics is potentially of broad significance. In both science and everyday life, we deal with a wide range of systems along the spectrum stretching from passive matter (and mechanisms made thereof) to the complex metacognitive capacities of adult human beings [18]. Especially interesting are the intermediate cases, such as cells, organs, swarms, artificial intelligences, autonomous vehicles, synthetic organisms, constructs made of active matter, and other increasingly-prevalent systems that have never existed in the evolutionary stream, all of which display a diverse range of capacities [10]. It is not enough to aim for minimizing estimates of agency (Morgan's Canon [107], often favored by scientists), because sub-optimal efficacy of prediction and control can result when we treat advanced reprogrammable systems as dumb machines and fail to appreciate their unexpected competencies. There are, of course, also downsides to over-estimating agency [43, 108-111]. The latter (false positives) result in low efficiency, while the former (false negatives) are associated with opportunity cost in engineering and regenerative medicine [49, 69, 112], not to mention serious ethical lapses; it is therefore both practically and ethically valuable to get our estimates of agency correct, not just low.

Indeed, this problem has only gotten thornier with time. During the early days of computers, we could accurately treat machines as passive (non-agentic) and humans as smart. However, with improvements in bioengineering and AI, it is becoming harder and harder to rely on such simple heuristics [113]. The field of Diverse Intelligence seeks principled frameworks for being able to recognize, predict, control, create, and ethically relate to a wide range of unconventional systems across the intelligence spectrum [18, 30].

Classical thinkers such as William James [1] were prescient enough to define intelligence in a cybernetic way, not tied to specific hardware (e.g., brains) or evolutionary origins. Definitions such as "competency to reach the same goal by different means" provide a sufficiently general, but empirically testable, way to define a spectrum of cognitive capacities. This drives hopes of being able to infer the design principles by which cognitive systems of different levels can be built, by creating novel active agents and studying the rich examples provided for us across the web of life [114, 115]. We strongly support [116-118] the definition of intelligence as problem-solving, utilizing an objectively observable 3rd person perspective in order to distinguish these highly tractable questions from the thorny debates of 1st-person consciousness [119-121].

Minimal chemical systems such as as active matter are being explored [122-130], as are basal competencies of "lower" organisms [85, 114, 131-138]. Here, we sought to produce an even more minimal, digital version of a system in which baseline expectation would not normally suggest any degree of intelligence. One advantage over biology, even at its lowest scales, is simplicity and transparency that guarantees that observed competencies of these algorithms are not due to an as-yet undiscovered explicit mechanism. Our goal is to show how an empirical stance, in which we use experimental tools of behavioral science and other disciplines to determine (rather than to make assumptions about) the level of cognition in a system, leads to interesting novel findings and discovery of capabilities we did not expect in a small, well-defined, deterministic, fully-transparent system created entirely by us. Prior examples of this approach include dynamical systems such as models of gene-regulatory networks, which exhibit not only complexity but several types of unexpected learning capacity [31, 32, 139-150]. We suggest that the discovery of unexpected problem-solving competencies (such as Delayed Gratification and segregation) that are not apparent from the component policies and algorithms themselves is a critical research program. The impacts of this effort, a central component of the emerging field of Diverse Intelligence, will have implications ranging across evolutionary developmental biology, philosophy of mind, AI alignment, and human flourishing via the safety of engineered composite systems.

**Acknowledgements:**

We thank David Ackley, Wesley Clawson, Franz Kuchling, and Julia Poirier for helpful comments on the project and manuscript. M.L. gratefully acknowledges support of the John Templeton Foundation via grant 62212. A.G. gratefully acknowledges the support of Astonishing Labs.

**Figure Legends**

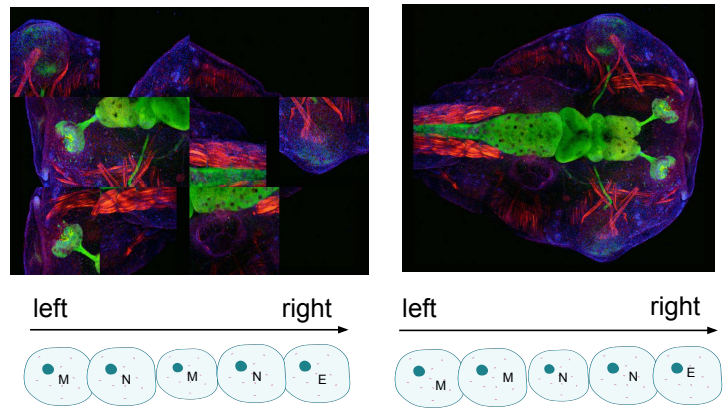

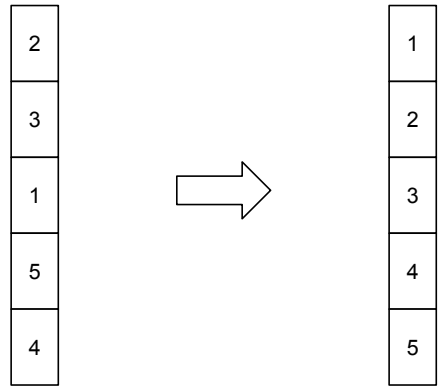

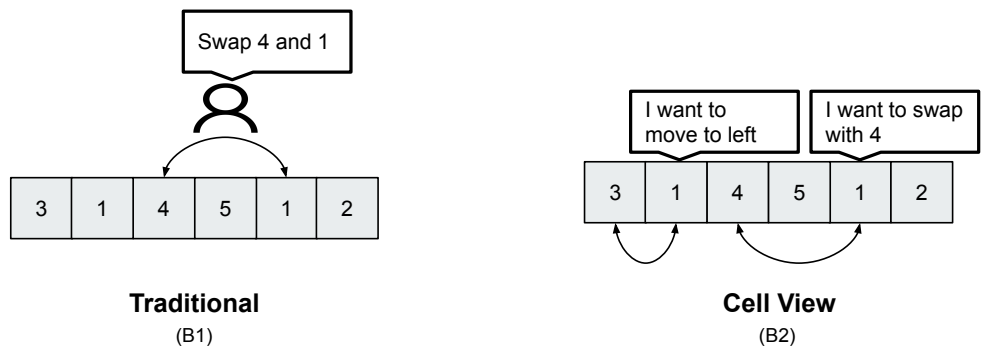

**Figure 1. Morphogenesis modeled as a sorting process.** In our abstraction, each element of the array was a cell, possessing an integer value that sets its final relative position along a 1-dimensional axis (simulating the correct target morphology that regulative development attempts

to implement along the anterior-posterior axis). Biologically, this corresponds to the tissue identity (eye, brain, nostril, etc.) that determines which neighbors each type of tissue expects to find, and thus set the stop condition for remodeling once those neighbor relations are attained. (A) The process of morphogenesis during development, repair, and metamorphosis included cells relocating to positions at which movement would cease (as determined for a given species) [151]. Here, 'E' = cells which will form the eyes; 'N' = cells which will form the nose; M = mouth cells. (B1) Traditional sorting algorithm performs as a top-down controller that makes decisions about the actions of the elements. (B2) Our cell-view version of this algorithm delegates the decision directly to each element, and the elements act based on their embedded logic and local conditions. Tadpole head organ photo used with permission from Dr. Helen Rankin, https://www.nikonsmallworld.com/galleries/2015-photomicrography-competition/transgenic-xenopus-laevis-tadpole-head-expressing-green-neurons.

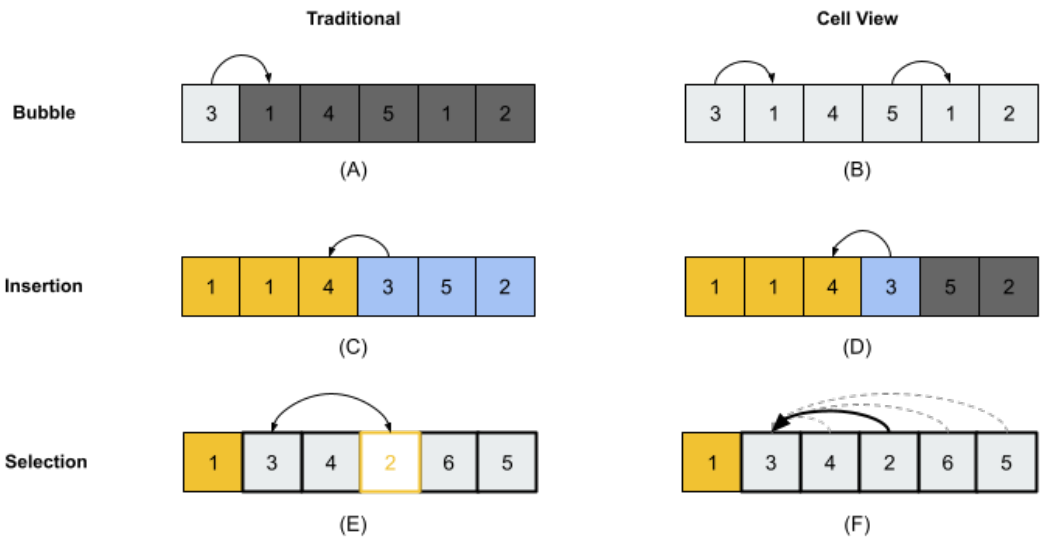

**Figure 2. Traditional sorting algorithms implemented as bottom-up drivers for cell behavior.** In conventional sorting algorithms, a single top-down controller implements a set of rules to move cells around. We sought to study the sorting process in a more biologically-inspired architecture, where each cell is a competent agent implementing local policies. We thus defined three bottom-up versions of common sort algorithms, where actions take place due to cells 'perspective (view) of their environment within the array. Shown here are examples of how cells are moved (traditional sort) or move themselves (cell-view sort) in each case. All cells have a chance to move at each time step (parallel). (A) In the traditional Bubble sort, the top-down controller chooses the first element on the left that is bigger than its right neighbor and keeps swapping it to the right until the right neighbor has larger or similar value, then repeats that process until all elements are in order. For the cell-view Bubble sort, all cells (elements) run in parallel and can compare their value with their neighbors', and decide to swap to the left or right based on that value comparison. The sort is completed when no cell can move. (B) In the traditional Insertion sort, the controller splits the array into two parts: sorted and unsorted. At the beginning, only the first element on the left is considered "sorted" and all remaining elements in the array are considered "unsorted". For each step, the controller chose the left-most element from the unsorted part, and then keeps swapping it to the correct place in the sorted part of the array. In the cell-view Insertion sort, each cell knows all cells on its left, and starts to swap with its left neighbor (if its value is smaller) when all cells to its right are sorted. (C) In the traditional Selection sort, the controller finds the smallest element (for an increase sort) from unvisited elements and puts it into the next position of the sorted part of the array for each step. In the cell-view Selection sort, every cell tries to swap to its own ideal target position. The swap will be denied if the current cell at that position has a smaller value, and then the ideal position of the cell shifts the right of the original ideal position.

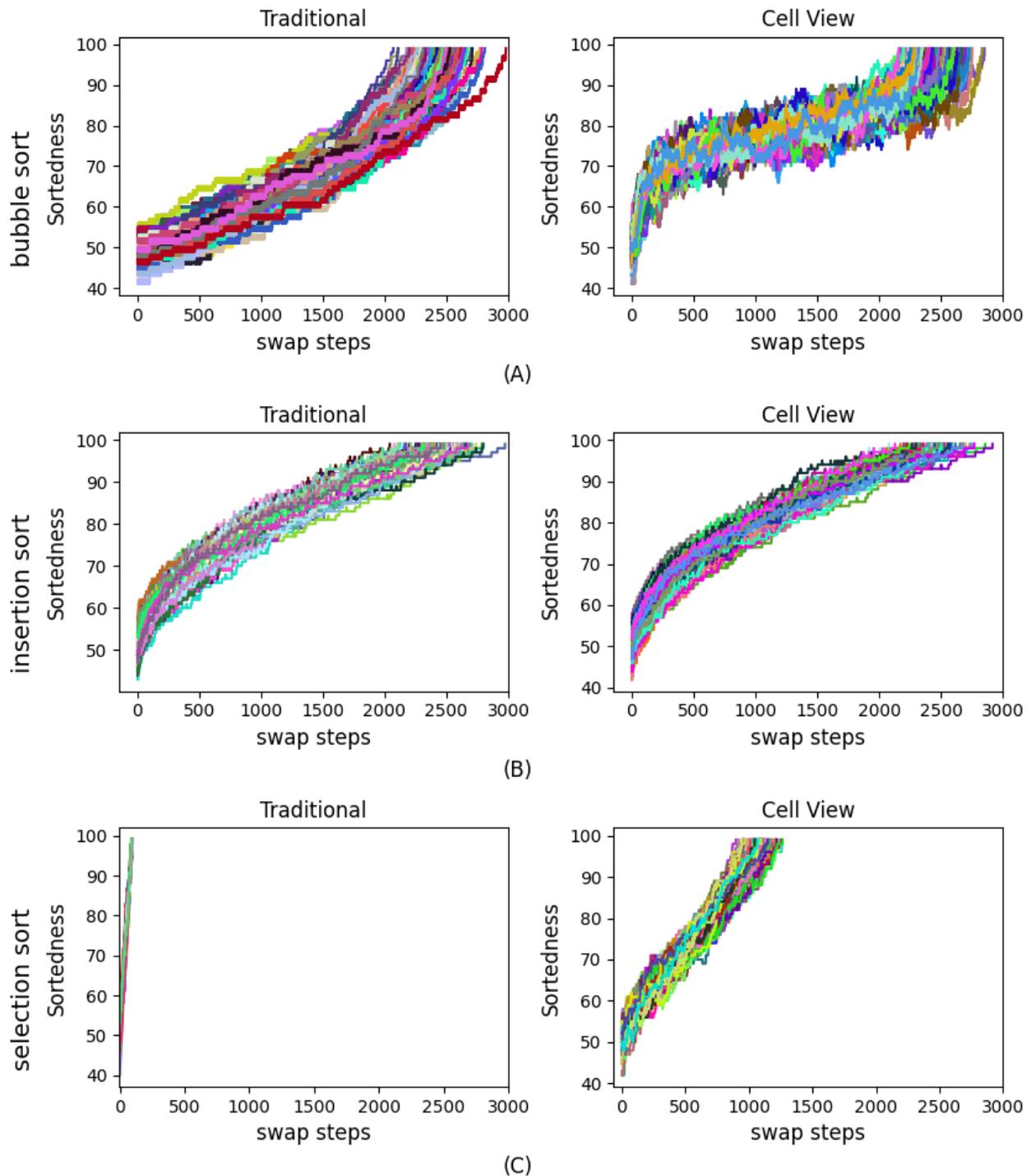

**Figure 3. Visualizing the sorting process as movement through sequence space.** Much as biological morphogenesis can be described as a trajectory through anatomical morphospace, here we view the progressive sorting process as the ability of traditional or cell-view sorting algorithms to navigate the state space of sequences toward the eventual goal of monotonicity. We define the degree of sequential order in the array of data, at any given time, as its *Sortedness*, here plotted on the Y axis. Each plot indicates the trajectory of 100 repeated experiments (sorting process on a random input number sequence, with no repeat digits). Just as their traditional counterparts do, cell-view sorts successfully completed the sorting process, navigating from a random state to the 100% fully sorted state. (A) shows the comparison of Sortedness change during the sorting process between traditional Bubble sort and cell-view Bubble sort. (B) shows the comparison of Sortedness change during the sorting process between traditional Insertion sort and cell-view Insertion sort. The difference between the two graphs is relatively small, because the implementation of cell-view Insertion sort always keeps the left side of the array sorted and allows one cell to join the sorted side each time which is very similar to the traditional Insertion sorting algorithm. (C) shows the comparison of Sortedness change during the sorting process between traditional Selection sort and cell-view Selection sort. The major difference between these two graphs is that the cell-view sorting process needs more swaps to complete the sort, because every cell can move to its ideal position and be swapped away when another cell with smaller value has the same ideal position.

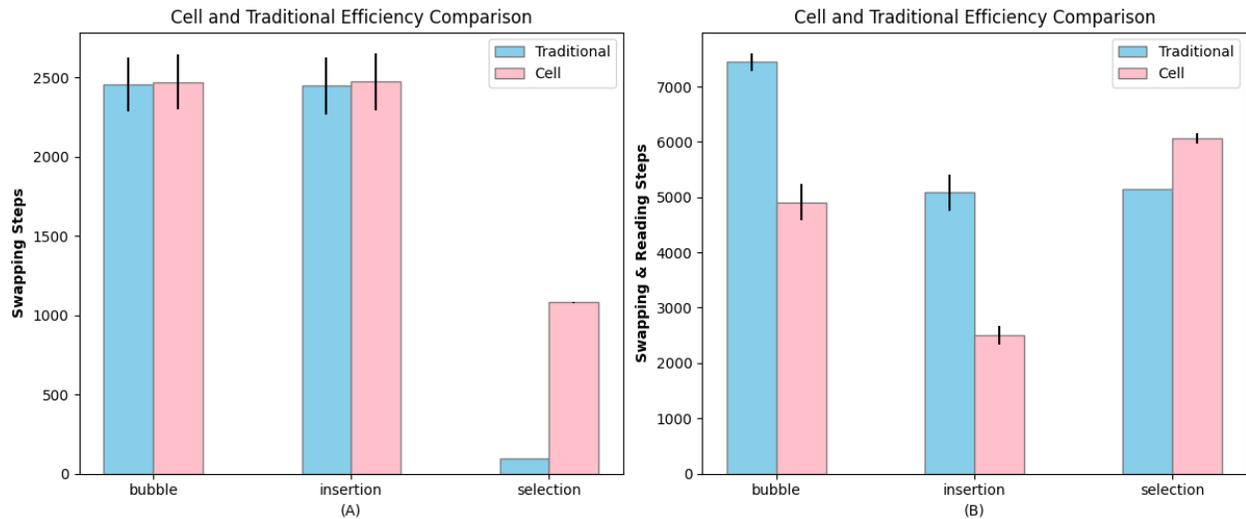

**Figure 4. Efficiency Comparison between Traditional and Cell-View Sort.** 100 experimental repeats were performed in three kinds of sort, in traditional and cell-view modes, to compare efficiencies of each method. When comparing only the active moves taken (A), it is seen that cell-view sort is almost exactly as efficient as the traditional version for Bubble and Insertion sort($p=0.24$), but is less efficient for Selection sort (the Z-test statistical value is 120.43, and p-value is 0). When comparing comparisons as well as moves (corresponding to the biological cost of sensing, as well as acting), the cell-view versions are actually *more* efficient for Bubble and Insertion Sorts (Z-test statistical values for bubble and insertion sort were -68.96 and -71.19 respectively, $p<<0.01$), while less efficient ($z=106.55$, $p<<0.01$) for Selection sort (B).

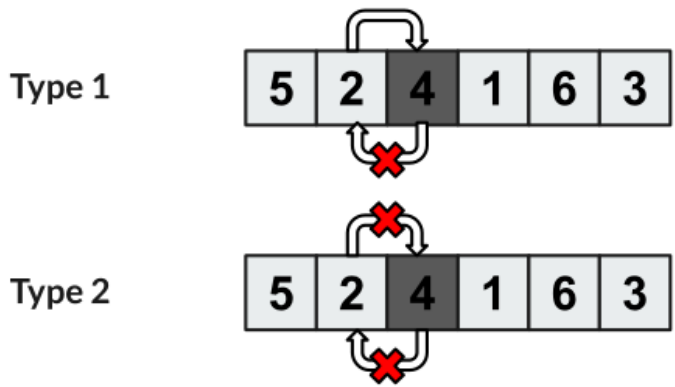

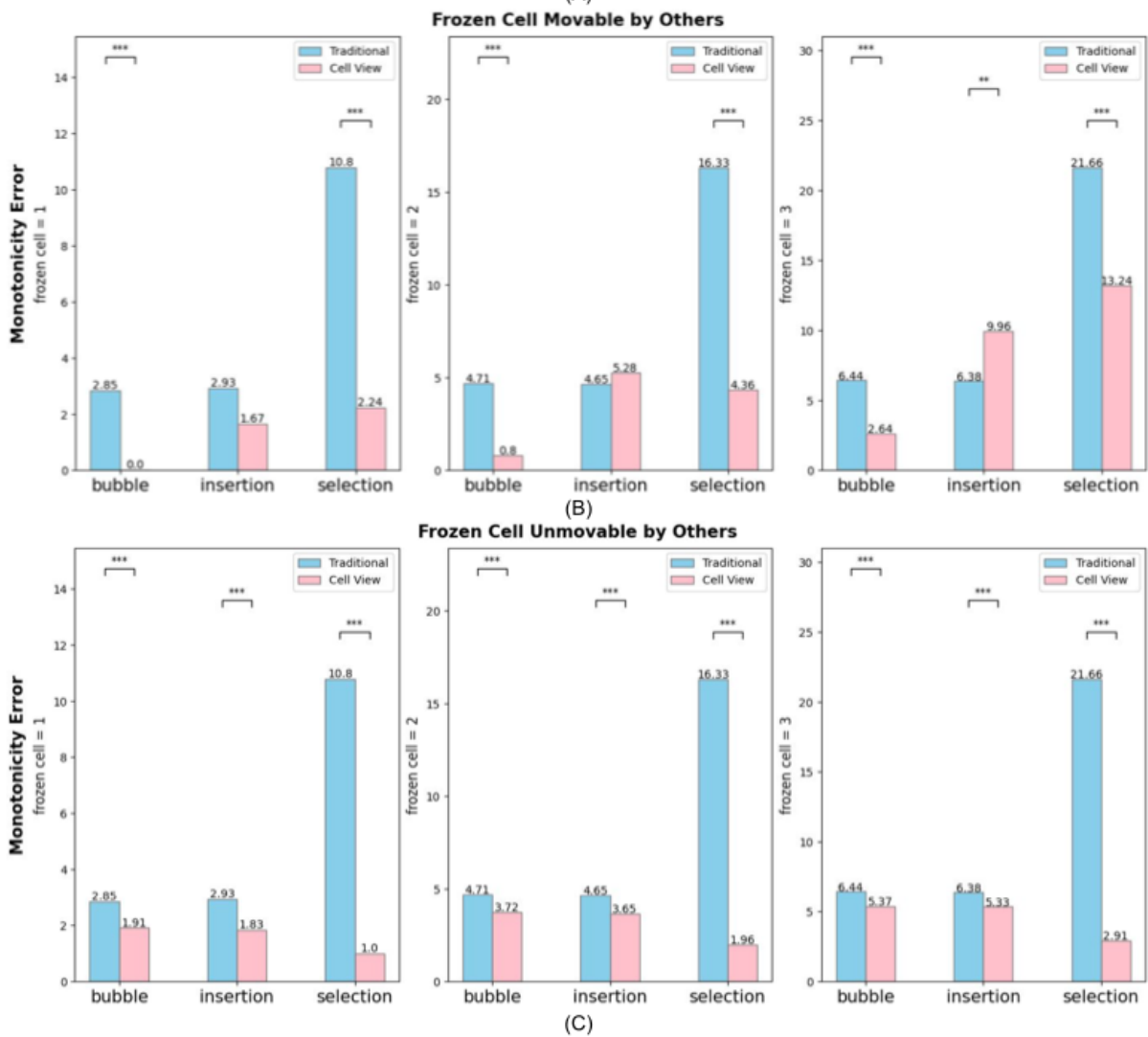

**Figure 5. Cell-View algorithms have better sorting performance for sorting with Frozen Cells.** Breaking the assumption of reliable media in traditional sorting algorithms enables exploration of emergent features of each algorithm in the face of errors: cells that need to move

but cannot. We tested two kinds of "defects" (A): cells that can be moved by others but cannot initiate any swaps ("lack of initiative"), and cells that are completely broken and neither initiate nor participate in swaps initiated by others ("lack of motility"). (B) shows when the Frozen Cell can be moved by others, the cell-view version of the sorting algorithms have less monotonicity error than the traditional version; i.e. the cell-view version has higher robustness (error tolerance). The same graph also shows that among cell-view Bubble sort, Insertion sort, and Selection sort, the Bubble sort has the highest error tolerance, with the Insertion sort having next-highest and the Selection sort lowest. (C) shows when the Frozen Cell is completely fixed, the cell-view version of the sorting algorithms also has higher error tolerance than the traditional version. Here, the cell-view Selection sort has the highest error tolerance.

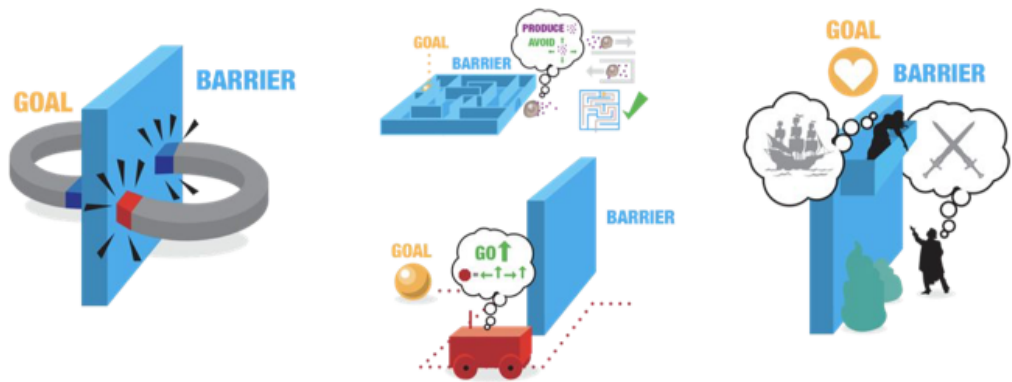

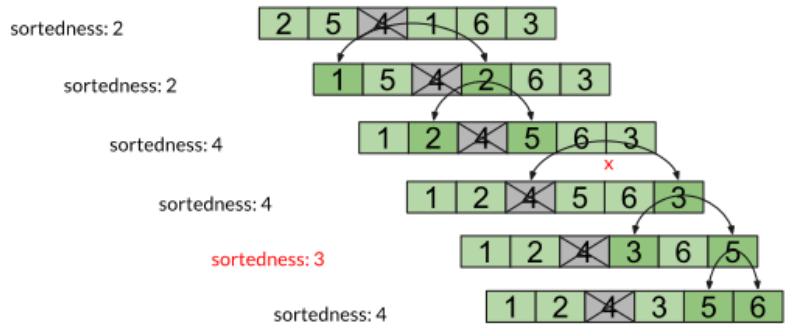

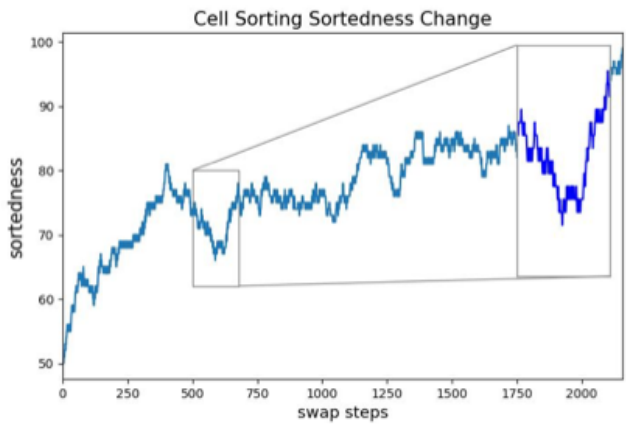 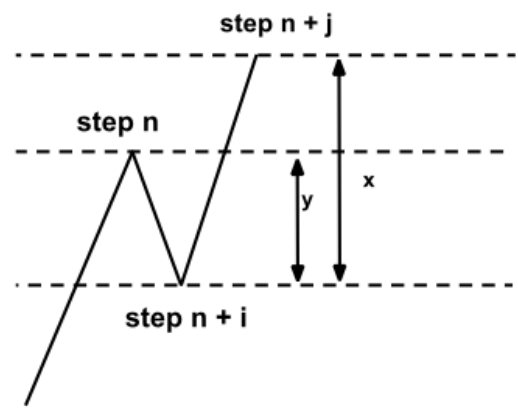

**Figure 6. Characterization of delayed gratification during the sorting process.** The ability of any agent to navigate towards its goal may include the property of *Delayed Gratification:* being able to temporarily go further away from its goal to achieve gains later in the process. (A) An illustration of William James' example of intelligence expressed as the ability to back-track from one's goals, which is not seen in simple energy-minimizing systems like magnets that will never go around a barrier to get closer (left-most panel). By contrast, back-tracking is extensively seen in human systems which can do complex planning (such as Romeo and Juliet, right-most panel), and exists to intermediate degrees in cells, tissues, and autonomous vehicles (middle panel). This has been proposed as a key parameter for defining a generic notion of intelligence [1], and is seen in our self-sorting cellular agents able to go around Frozen Cells. Artwork in panel A is courtesy of Jeremy Guay of Peregrine Creative. (B) An example of going around a Frozen Cell (broken cell) using a short array, in which the cell with value 3 wants to swap to 3rd position but it can't, so it does the best possible alternative – it swaps to 4th position. This move temporarily decreases the Sortedness, until the cells with values 5 and 6 have swapped and caused Sortedness to increase again. (C) In the arrays with more Frozen Cells, there were multiple local reductions of Sortedness during the process. (D) Such back-tracking and the subsequently-realized gains can be used to define a *Delayed Gratification* index during the sorting process, which is calculated by using the total consecutive increasing value (x) of Sortedness after the drop as the numerator, and the total consecutive decreasing value (y) of Sortedness before the increase as the denominator.

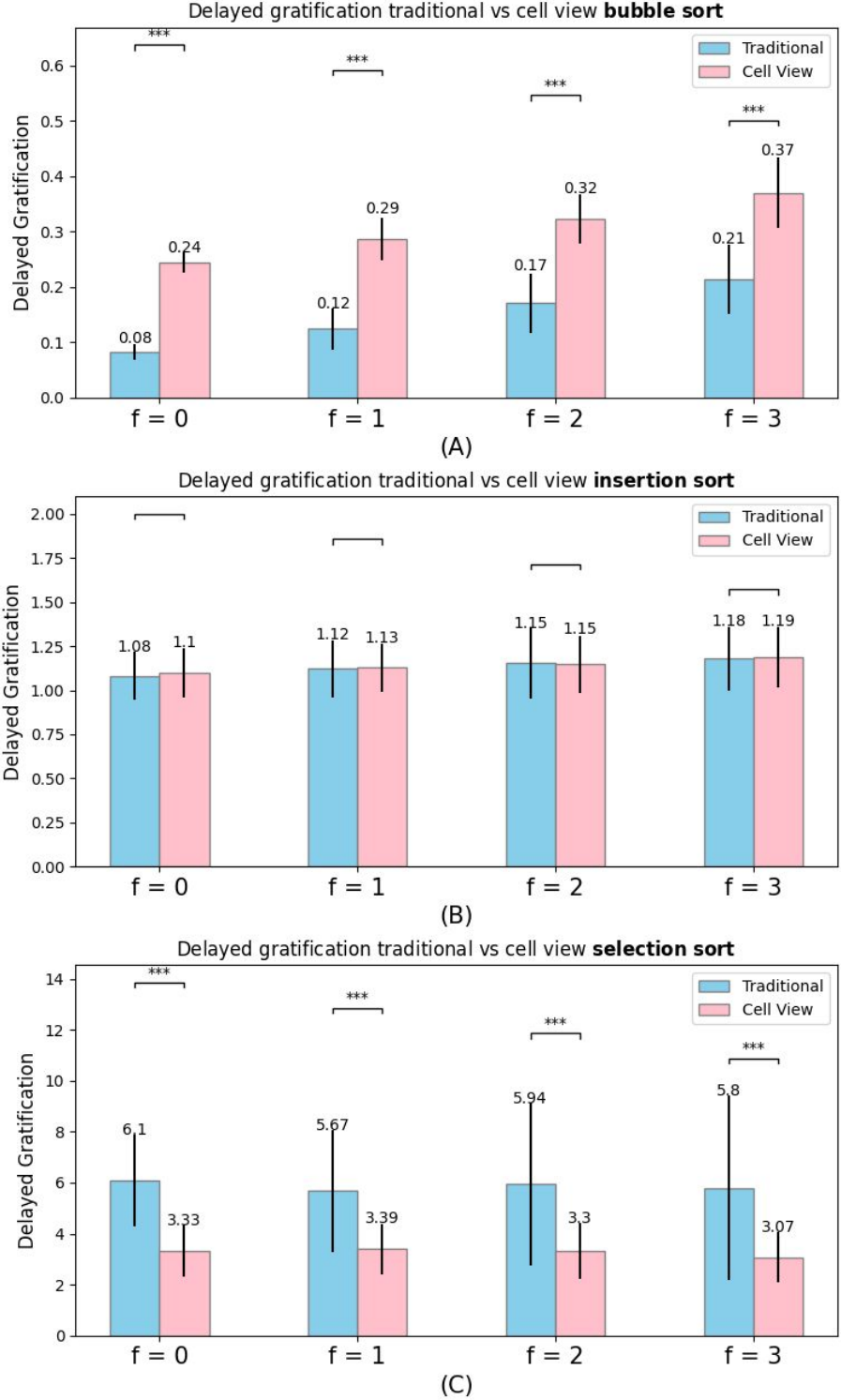

**Figure 7. Delayed Gratification Comparison between Traditional and Cell-View Sort.** Delayed Gratification is a characteristic shown in all the sorting algorithms for both traditional and cell-view, because all the experiments show some degree of Delayed Gratification. However, the Delayed Gratification is not merely evaluating the inefficient movements that randomly go backwards, because the figures reflect the increasing trend of Delayed Gratification as the number of Frozen Cells increases. (A) shows that the cell-view Bubble sort gets more Delayed Gratification than the traditional algorithm ($Z=34.04$, $p\ll0.01$). (B) shows that the cell-view Insertion sort gets slightly more Delayed Gratification than the traditional algorithm ($Z=0.60$, $p=0.55$). Panels (A) and (B) show that the Bubble and Insertion sorts get more Delayed Gratification as the number of Frozen Cells increases. Image (C) shows that the cell-view Selection sort has less Delayed Gratification than the traditional algorithm ($Z=-17.21$, $p\ll0.01$) and has no clear relation between Delayed Gratification and the number of Frozen Cells. Comparing (A), (B) and (C), the Selection sort does the best in Delayed Gratification ($Z=40.81$, $p\ll0.01$), and the Insertion sort does better than the Bubble sort for both traditional and cell-view algorithms ($Z=98.04$, $p\ll0.01$).

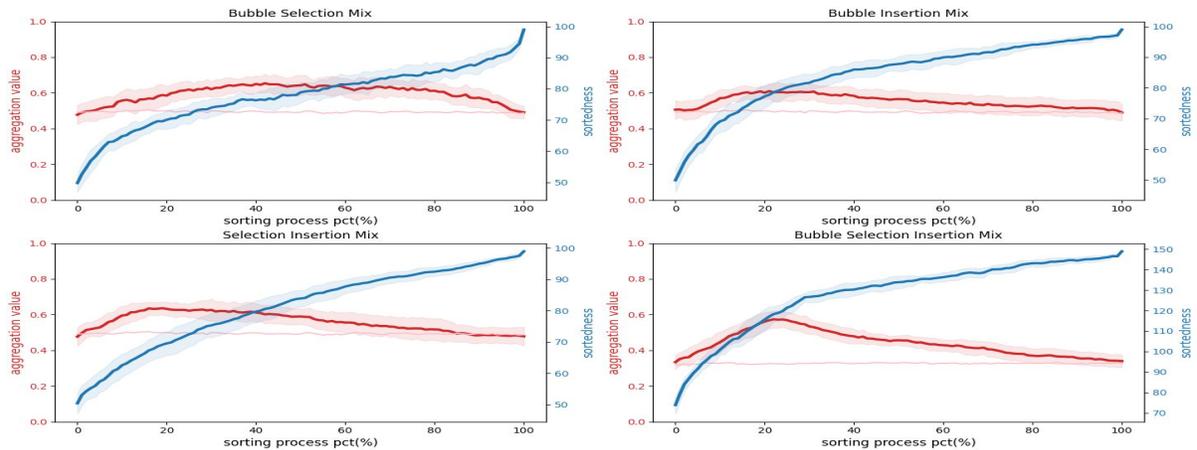

(A)

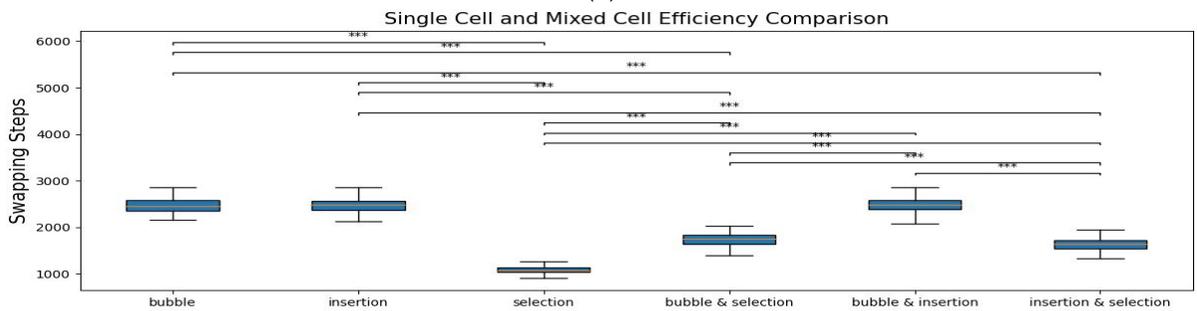

(B)

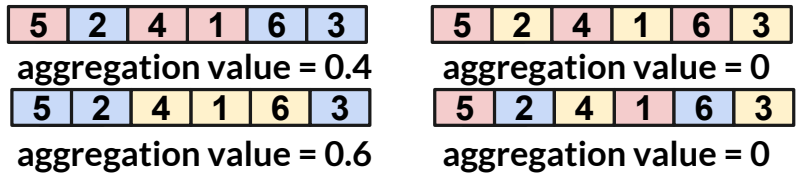

(C)

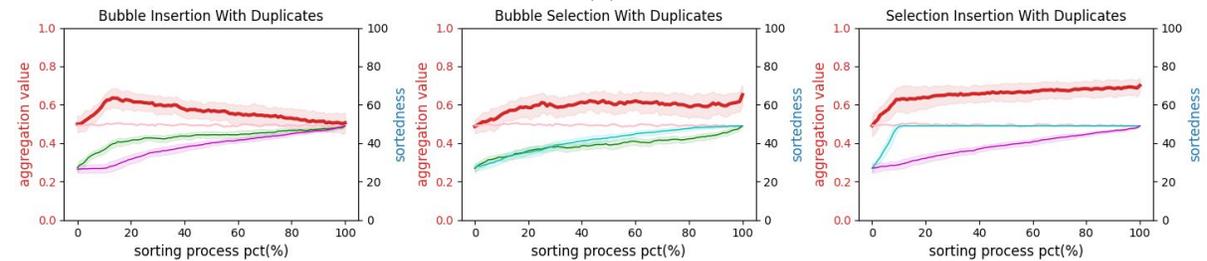

(D)

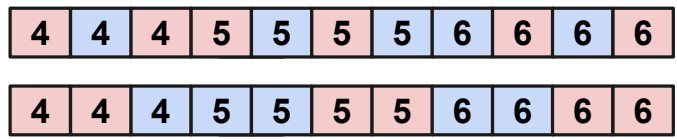

(E)

**Figure 8. Unexpected aggregation of Algotypes occurs in chimeric arrays.** A cell-view (local) implementation of sorting policies enables an experiment that is not done in traditional sorting: chimeric arrays in which individual cells follow their own distinct policies ("Algotype") for how to move (corresponding to chimeric embryos consisting of cells of 2 different lineages [9]);

efficiencies of such chimeric individuals are shown in Supplemental Panel (B). To study how such chimeric collectives behave, we investigated 100 repeats of scenarios where each combination of the 3 sort algorithms was represented equally; note that the algorithms were not modified in any way and thus do not have any provision for knowing their own Algotype or that of their neighbors. (C). To understand the relative spatial distribution of cells executing each algorithm within the array during the sorting process, we defined *aggregation value*: the probability that a cell's neighbor is of the same type as itself. Algotypes were assigned to each cell randomly in each experimental array and did not change during the course of the sort. Panel (A) shows the results of each possible combination of sort algorithms within an array. The blue lines indicate the progress of the sort itself (the Sortedness value). The pink lines indicate the Aggregation Value when two identical sorts are used - this negative control shows, as expected, that there is no significant deviation from 50% chance. The red line indicates the Aggregation Value of each kind of sort. As expected, at the beginning the Aggregation Value is 50%, since types are assigned to cells randomly. Likewise, at the end, the Aggregation Value is back to 50% since the array is sorted only by each cell's Value, with no regard for Algotype, and the Algotypes were randomly assigned. Remarkably, significant aggregation was observed during the sorting process, peaking at 60% (with peaks that occur at slightly different times during the sorting process for each of the chimeric combinations). (D) We then allowed duplicate digits in the arrays, so that some instances of each number would be of each of the types, in order to see what maximum aggregation could be achieved if the explicit (monotonic numbers) and implicit (aggregation of types) goals were made compatible. We observed that the final Aggregation Values in (D) were larger than the final Aggregation Values in (A).

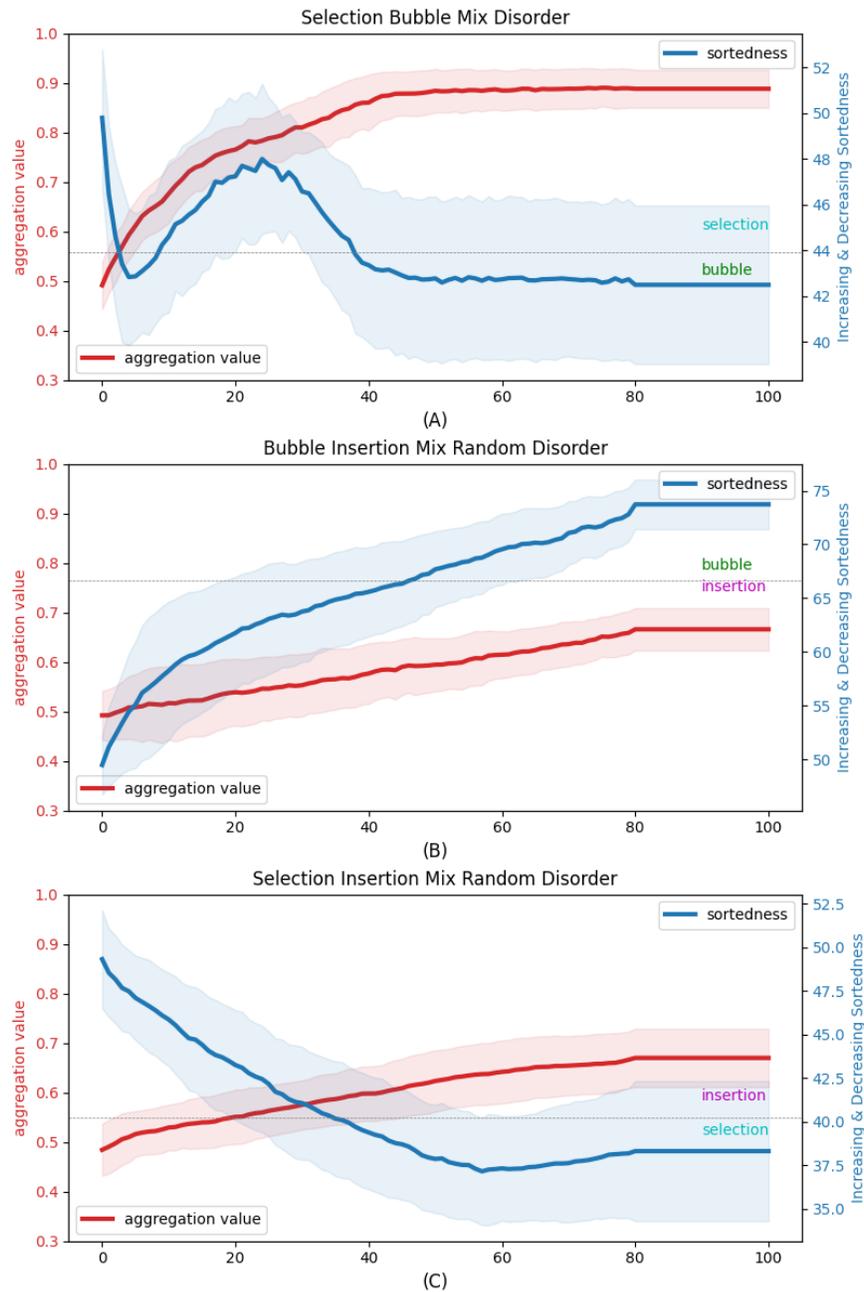

**Figure 9. Cell Aggregation for Cells with Different Sorting Directions.** In biological chimeras made from the cells of different cells, those cells not only use different policies for their activity, but they are actively trying to build different anatomical patterns. Thus, we asked what chimeric sorts would do if the two types of cells were trying to sort in opposite directions (monotonic increasing vs. decreasing). For each type of sort, there is a consistent but not linear dynamic of conflict between the 2 algorithms. After some initial back-and-forth, Sortedness flattens out (reaches a stable point after which nothing will change), with algorithms "winning" (being more effective than their competitors) in this order: Bubble > Selection > Insertion. The Aggregation Values also flatten out: the final Aggregation Values were all higher than the starting averages, which were ~0.5.